\documentclass[runningheads]{llncs}
\usepackage[T1]{fontenc}
\usepackage{graphicx}

\usepackage{epsfig}
\usepackage{amsmath}
\usepackage{amssymb}
\usepackage{subcaption}
\usepackage{booktabs}
\usepackage{braket}
\usepackage{xcolor}
\usepackage{float}
\usepackage{wrapfig}

\usepackage[linesnumbered,ruled,vlined]{algorithm2e}

\SetCommentSty{mycommfont}
\SetKwInput{KwInput}{Input}
\SetKwInput{KwOutput}{Output}

\begin{document}
%
% \title{Improving Disentangled Representation with Independence Constraint: A Perspective from Epistemology}

\title{Independence Constrained Disentangled Representation Learning from Epistemological Perspective}

\titlerunning{Independence Constrained Disentangled Representation Learning}
% If the paper title is too long for the running head, you can set
% an abbreviated paper title here
%
\author{Ruoyu Wang \inst{1} \and
Lina Yao \inst{1,2}}
% \author{Anonymous submission}
%
\authorrunning{Wang et al.}
% First names are abbreviated in the running head.
% If there are more than two authors, 'et al.' is used.
%
% \institute{Princeton University, Princeton NJ 08544, USA
% \email{lncs@springer.com}
% }
\institute{University of New South Wales \and Commonwealth Scientific and Industrial Research Organisation, Australia \email{ruoyu.wang5@unsw.edu.au}, \email{lina.yao@unsw.edu.au}}

\maketitle              % typeset the header of the contribution
\begin{abstract}
Disentangled Representation Learning aims to improve the explainability of deep learning methods by training a data encoder that identifies semantically meaningful latent variables in the data generation process. Nevertheless, there is no consensus regarding a universally accepted definition for the objective of disentangled representation learning. In particular, there is a considerable amount of discourse regarding whether should the latent variables be mutually independent or not. In this paper, we first investigate these arguments on the interrelationships between latent variables by establishing a conceptual bridge between Epistemology and Disentangled Representation Learning. Then, inspired by these interdisciplinary concepts, we introduce a two-level latent space framework to provide a general solution to the prior arguments on this issue. Finally, we propose a novel method for disentangled representation learning by employing an integration of mutual information constraint and independence constraint within the Generative Adversarial Network (GAN) framework. Experimental results demonstrate that our proposed method consistently outperforms baseline approaches in both quantitative and qualitative evaluations. The method exhibits strong performance across multiple commonly used metrics and demonstrates a great capability in disentangling various semantic factors, leading to an improved quality of controllable generation, which consequently benefits the explainability of the algorithm.

\keywords{Disentangled Representation Learning  \and Generative Adversarial Network \and Explainability}
\end{abstract}
\section{Introduction}
\label{intro}
Representation learning is widely recognized as a fundamental task in the field of machine learning, as the efficacy of machine learning methods heavily relies on the quality of data representation. It is suggested that an ideal representation should be disentangled \cite{bengio2013representation}, which means it can identify the genuine generative factors hidden in the observed data, and the latent variables should be semantically meaningful and correspond to the ground truth generative factors.

\begin{figure}
    \center
    \begin{subfigure}[b]{0.4\textwidth}
        \includegraphics[width=\linewidth]{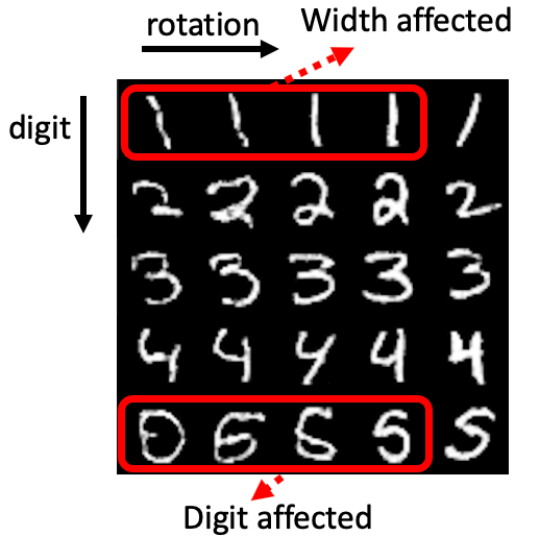}
        \caption{}
        \label{fig:mnist_samples_without}
    \end{subfigure}
    \begin{subfigure}[b]{0.4\textwidth}
        \includegraphics[width=\linewidth]{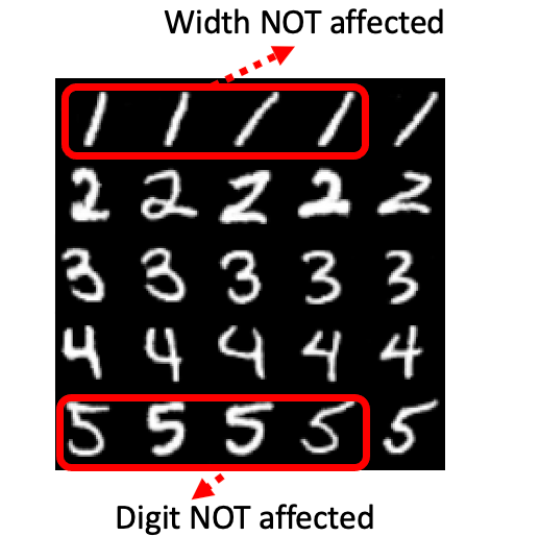}
        \caption{}
        \label{fig:mnist_samples_with}
    \end{subfigure}
  \caption{Disentangled Representation Learning learns semantically meaningful latent variables such as \textit{Rotation} and \textit{Digit}. Our method outperforms existing methods by encouraging the factors to be further separated. For example, on MNIST: (a) Without the independence constraint in our method, the \textit{digit} and \textit{width} are affected when traversing on variable \textit{rotation}; (b) With the independence constraint, \textit{digit} and \textit{width} are NOT affected when traversing on \textit{rotation}.}
  \label{fig:mnist_samples}
\end{figure}

However, there is no general agreement on a formal definition of disentangled representation \cite{bengio2013representation} \cite{higgins2018towards} \cite{suter2019robustly}. Despite the lack of agreement on the formal definition of disentangled representation learning, existing methods suggested that two quantities are significant in disentangled representation learning: 1) Mutual Information between the latent variables and the data \cite{chen2016infogan}; and 2) Independence between the latent variables \cite{kim2018disentangling}\cite{chen2018isolating}.

Nevertheless, regarding the second quantity \textit{Independence between the latent variables}, it is worth noting that there is a lack of consensus about whether latent variables should be mutually independent in disentangled representation learning. While some \cite{bengio2013representation,higgins2018towards,higgins2017beta,kim2018disentangling,chen2018isolating} suggest that hidden factors should be strictly independent, some other works \cite{suter2019robustly,reddy2022causally,yang2021causalvae,shen2022weakly} argued that causal relationships exist between generative factors, thus they are not necessarily independent. Therefore, these arguments lead us to ask:

\begin{center}
    \textit{What should be considered as generative factors? What should be independent in latent space? And what should be causally connected?}
\end{center}

To answer these questions, it is crucial to understand how humans perceive and comprehend these factors and their relationships, because the fundamental objective of disentangled representation learning is to extract factors that are \textbf{interpretable to us human}. Therefore, in this paper, we first answer the above questions by borrowing the concepts from epistemology. Then, based on these interdisciplinary theories, we introduce a unified framework to consolidate prior arguments regarding the relationships between latent variables. Finally, after clarifying these questions, we propose a novel method for disentangled representation learning that jointly optimizes the two objectives mentioned earlier: 1) the mutual information objective and 2) the independent objective. The contribution of this paper is threefold:

\begin{itemize}
  \item We establish a conceptual bridge between epistemology and disentangled representation learning to facilitate the understanding of the data generation process and disentangled representation learning.
  \item We introduce a two-level latent space framework to unify the prior arguments regarding the relationships between generative factors and latent variables in disentangled representation learning.
  \item We propose a novel method for disentangled representation learning to jointly optimize the mutual information and the independent objectives, which outperforms the baseline methods consistently on multiple evaluation metrics.
\end{itemize}

\section{Our Method}
\subsection{A Perspective of Epistemology}
\label{method_epistemology}

As discussed in Section~\ref{intro}, comprehending the relationships between latent variables in disentangled representation learning necessitates an understanding of how humans perceive these factors. Because the concept of \textit{interpretability} inherently implies interpretability to humans; therefore, interdisciplinary insights from epistemology are indispensable in this context.

\begin{figure*}
     \centering
     \begin{subfigure}[b]{0.49\textwidth}
         \centering
         \includegraphics[width=\textwidth]{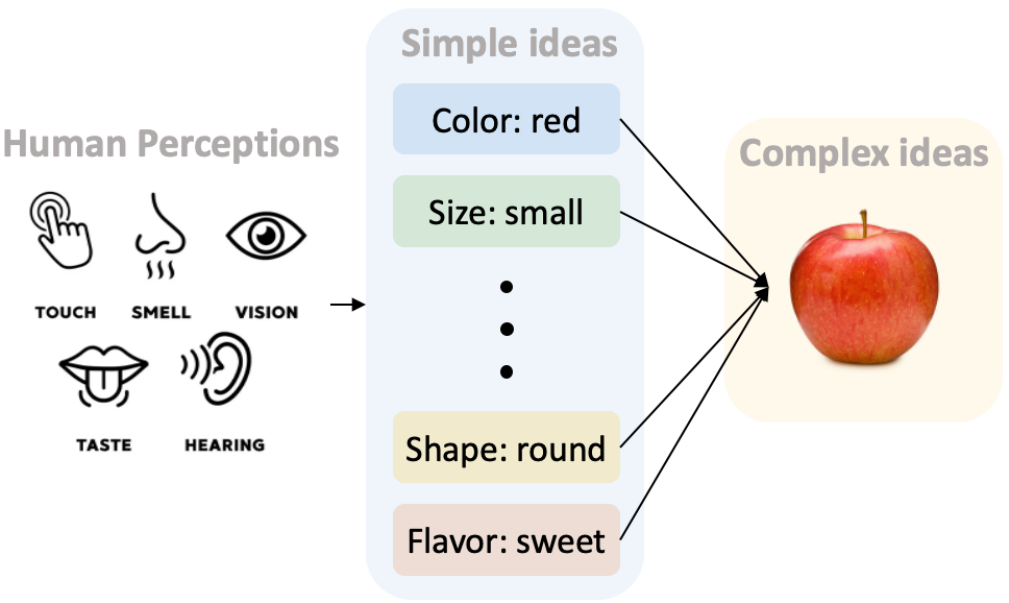}
         \caption{Ideas in Epistemology}
         \label{fig:epistemology}
     \end{subfigure}
     \hfill
     \begin{subfigure}[b]{0.49\textwidth}
         \centering
         \includegraphics[width=\textwidth]{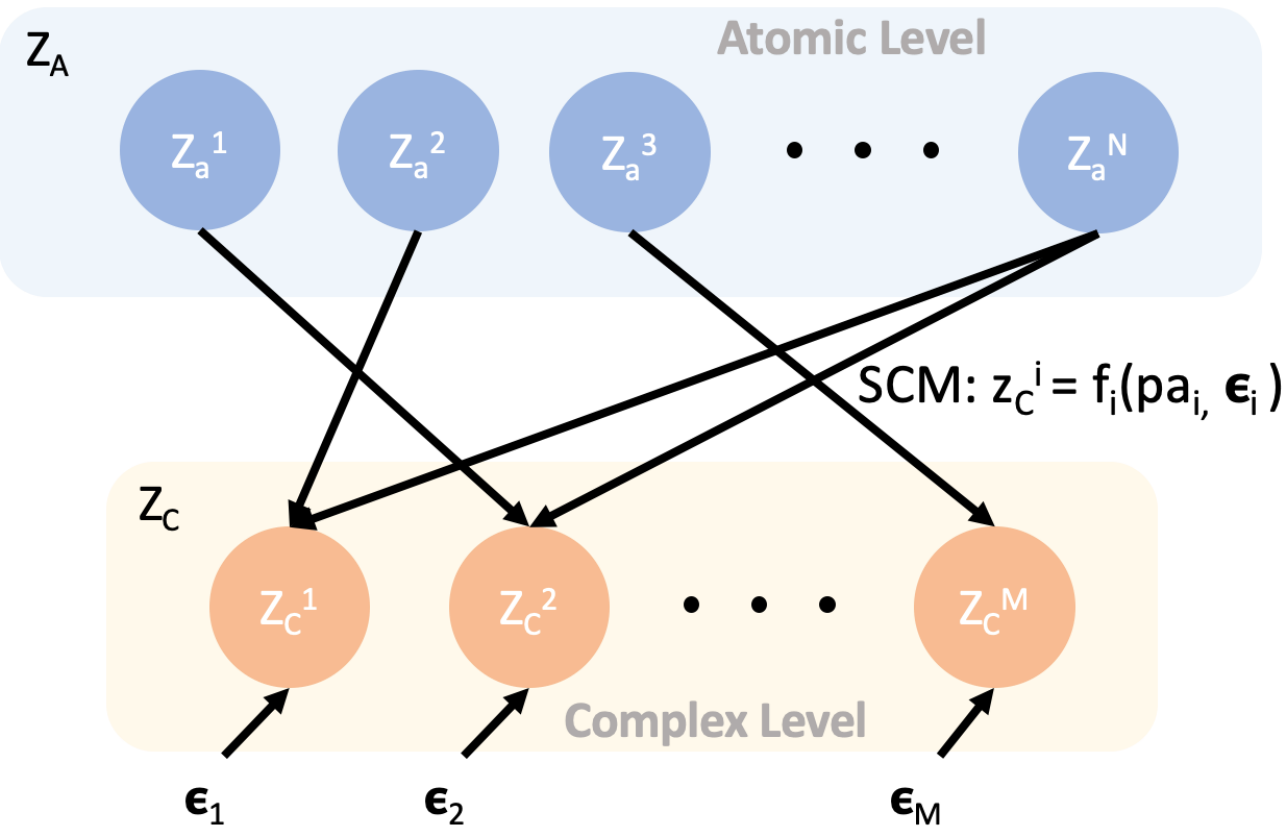}
         \caption{The two-level latent space framework}
         \label{fig:two_level_framework}
     \end{subfigure}
    \caption{(a) In epistemology, the complex idea \textit{apple} is composed by simple ideas such as its \textit{colour}, \textit{shape} and \textit{size}, which are irreducible and mutually independent; (b) Our proposed framework groups latent variables into two levels: 1) Atomic Level $Z_{A}$: comprises factors that are basic and irreducible such as colour and shape of an apple, where all factors are mutually independent; 2) Complex Level $Z_{C}$: comprises factors that are derived from the atomic level, such as the concept of apple. The two levels are connected by causal relationships.}
    \label{fig:method}
\end{figure*}

In epistemology, mental representations of perceptions are called {\it ideas}, which can be grouped into two categories, simple ideas and complex ideas \cite{hume2003treatise}. Simple ideas are basic, indivisible concepts that form the foundation of our knowledge, and complex ideas are more advanced concepts that are built upon multiple simple ideas. For instance, when we imagine an apple (Figure~\ref{fig:epistemology}), the sensory perceptions of an apple such as its colour, shape and taste are irreducible, and thus are regarded as simple ideas. These simple ideas can then be combined to form the complex idea of an apple as a whole. 

Taking these theories as a reference, we argue that the disagreement regarding whether latent variables should be independent arises from the lack of understanding of this hierarchical structure of human concepts. Existing works instinctively consider ALL latent variables in a single latent space. However, building upon these interdisciplinary theories, we contend that interpretable latent variables in disentangled representation learning should also follow a hierarchical structure in a similar way as demonstrated in Figure~\ref{fig:epistemology}.

\subsection{Two-level Latent Space Framework}
\label{two_level_framework}
Inspired by the interdisciplinary theories introduced in Section~\ref{method_epistemology}, we propose a two-level latent space framework to consolidate prior arguments on the interrelationships between the latent variables, as illustrated in Figure~\ref{fig:two_level_framework}. 

The framework groups all latent variables into two levels: 1) \textbf{Atomic Level}: which corresponds to the simple idea in epistemology, and comprises the factors that can be directly perceived from the observed data, and latent variables in this level should be \textbf{mutually independent}. 2) \textbf{Complex Level}: which comprises concepts derived from the atomic level variables. The two levels are connected by causal relationships and can be modelled by the Structural Causal Model (SCM). Within this framework, atomic-level latent variables should be \textbf{mutually independent}, and complex-level variables are \textbf{not necessarily independent} because the atomic-level variables may work as confounders due to the causal relationships between the two levels. 

We argue that prior arguments regarding the relationships between latent variables are primarily due to their focus on a set of selected factors in certain datasets, thus having different views on this issue. In contrast, our framework, supported by the theories in epistemology\cite{hume2003treatise} and also cognitive science \cite{murphy2004big,rips1973semantic,berlin2014ethnobiological}, offers a general explanation that is adaptable to various scenarios, and consolidates prior arguments regarding the interrelationships between latent variables. In this paper, we concentrate on the datasets that consist only of atomic-level latent variables, thus all latent variables are independent. Therefore, we leverage the constraints of independent and mutual information to augment the efficacy of our proposed method for disentangled representation learning.

\subsection{Method Formulation}
\begin{figure*}[t]
    \centering
    \begin{subfigure}[b]{1\textwidth}
        \includegraphics[width=\linewidth]{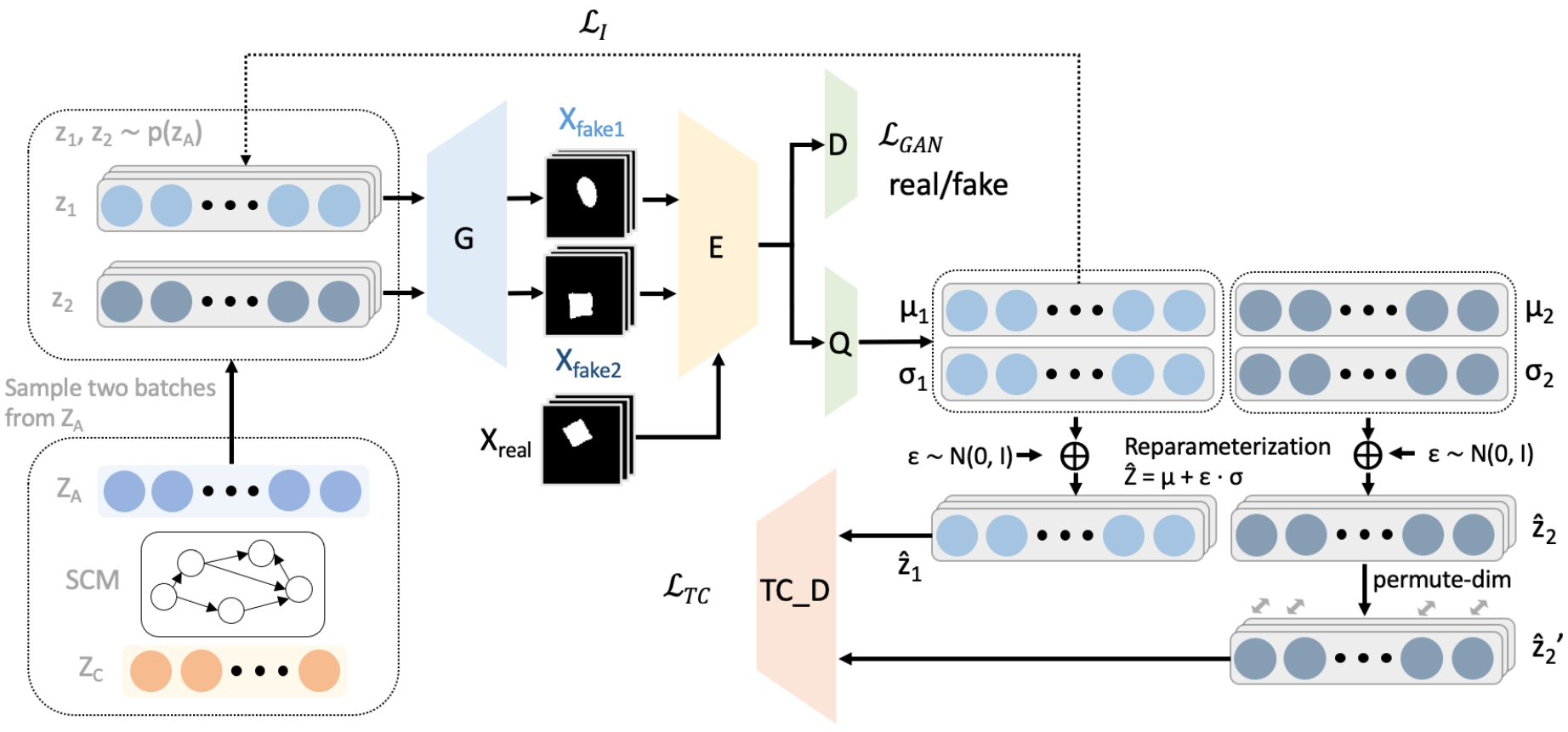}
        \label{fig:tcgan_drt}
    \end{subfigure}
  \setlength{\abovecaptionskip}{-5pt}
  \caption{Framework of TC-GAN. Two batches of latent variables $z_{1}, z_{2}$ are sampled from the atomic level latent space and passed to the generator $G$ to generate the fake images. The discriminator $D$ and the auxiliary network $Q$ share the same data encoder $E$. The network $Q$ outputs the predicted mean $\mu_{1}, \mu_{2}$ and variance $\sigma_{1}, \sigma_{2}$ of the latent factor. Then, we employ the re-parameterization trick and the permute-dim algorithm to aid the $TCD$ in estimating Total Correlation.}
  \label{fig:tcgan_drt_framework}
  % \vspace{-10pt}
\end{figure*}

After clarifying the legitimacy of applying independence constraints in this problem in Section~\ref{two_level_framework}, we introduce our proposed method in this section. We build our method in the paradigm of Generative Adversarial Networks (GAN). Formally, GAN solves the minimax optimization problem (Equation~\ref{eq:loss_gan}) by utilizing a generator $G$ and a discriminator $D$, and the generator $G$ could learn the real data distribution $P_{real}(x)$ when this framework converges. 
\begin{equation}
    \min_{G}\max_{D} \mathcal{L}_{GAN}(D,G) = \mathbb{E}_{x \sim P_{real}} \left[ \log D(x) \right] + \mathbb{E}_{z \sim noise}  \left[ \log (1-D(G(z))) \right]
  \label{eq:loss_gan}
\end{equation}

However, this framework imposes no restrictions on the semantic meaning of the latent variable $z_{i}$. To encourage the latent variables to possess semantic meanings, as introduced in Section~\ref{intro}, we apply the two constraints in this framework: 1) Mutual information between the latent variables and the data \cite{chen2016infogan}; and 2) Independence between the latent variables \cite{kim2018disentangling,chen2018isolating}. 

For the constraint on mutual information, we adapt the implementation of InfoGAN \cite{chen2016infogan}. First, the latent variables are decomposed into latent code $z$ which controls the semantics in the image, and noise $\epsilon$ which is considered incompressible. Then, an auxiliary network Q is introduced into the framework to maximize the lower bound of mutual information between the latent code $z$ and the image (Equation~\ref{eq:mi_elbo}). In practice, a regularization term $\mathcal{L}_{I}(G,Q)$ (Equation~\ref{eq:loss_miloss}) is introduced into the framework to maximize the Mutual Information between latent variable $z$ and the generated data.
\begin{equation}
    I(z,G(z,\epsilon)) \geq \mathbb{E}_{x \sim G(z,\epsilon)} \left[ \mathbb{E}_{z' \sim P(z|x)} \log Q(z'|x) \right] + H(z)
  \label{eq:mi_elbo}
\end{equation}
On the other hand, for the independence constraint, we aim to minimize the \textit{Total Correlation} \cite{watanabe1960information} between latent variables, and apply this constraint to the learned latent code $\hat{z}$ such that:
\begin{equation}
    TC(\hat{z}) = KL(q(\hat{z})|| \prod_{i}q(\hat{z}_{i}))
  \label{eq:tc_constraint}
\end{equation}
where $\hat{z}$ is the samples drawn from the learned posterior distribution obtained by $Q(G(z))$. Therefore, by integrating these two constraints, the overall objective of our method becomes:

\begin{equation}
    \min_{G,Q}\max_{D} \mathcal{L}_{TCGAN} = \mathcal{L}_{GAN}(D,G) - \lambda \mathcal{L}_{I}(G,Q) + \beta \mathcal{L}_{TC}(G,Q)
  \label{eq:loss_infotcgan}
\end{equation}
where
\begin{equation}
    \mathcal{L}_{I}(G,Q) = \mathbb{E}_{z \sim P_{(z)}, x \sim G_{(z,\epsilon)}}\left[ \log Q(z|x) \right]
  \label{eq:loss_miloss}
\end{equation}
\begin{equation}
    \mathcal{L}_{TC}(G,Q) = KL(q(\hat{z})|| \prod_{i}q(\hat{z}_{i}))
  \label{eq:loss_tcloss}
\end{equation}

Thus far, we have formulated our method by integrating the two constraints in the paradigm of Generative Adversarial Network. However, the total correlation term (Equation~\ref{eq:loss_tcloss}) is intractable. In Section~\ref{method_TC_estimate}, we introduce the method to estimate this term and how this estimation process is integrated into our framework, then give an end-to-end illustration of our method.

\subsection{Total Correlation Estimation}
\label{method_TC_estimate}

To estimate the Total Correlation term (Equation~\ref{eq:loss_tcloss}), we adapt the method from FactorVAE \cite{kim2018disentangling}. Specifically, we utilize the Density-Ratio trick, which trains a Total Correlation Discriminator $TCD$ to predict the probability that the input vector is from $q(\hat{z})$ rather than $\prod_{i}q(\hat{z}_{i})$, and then the Total Correlation term can be estimated by Equation~\ref{eq:density_ratio_trick}.

\begin{equation}
\label{eq:density_ratio_trick}
TC(\hat{z}) = \mathbb{E}_{q(\hat{z})}  \left[ \log \frac{q(\hat{z})}{\prod_{i}q(\hat{z}_{i})} \right] \approx \mathbb{E}_{q(z)} \left[ \log \frac{TCD(\hat{z})}{1-TCD(\hat{z})}\right]
% = \mathbb{E}_{q(\hat{z})} \left[ \log q(\hat{z}) \right] - \mathbb{E}_{q(\hat{z})} \left[\log \prod_{i}q(\hat{z}_{i}) \right]
\end{equation}

The end-to-end framework of our method is illustrated in Figure~\ref{fig:tcgan_drt_framework}. On top of the vanilla GAN framework which comprises a generator $G$, a data encoder $E$ and a Discriminator head $D$, we first utilize an auxiliary network $Q$ to predict the mean and variance of the latent variables of the generated data, as implemented in InfoGAN \cite{chen2016infogan}. Then, we sample $\hat{z}$ by utilizing a reparametrization trick to ensure the differentiability of the framework. By employing this approach, we acquired samples from the distribution $q(\hat{z})$. On the other hand, samples from $\prod_{i}q(\hat{z}_{i})$ cannot be sampled directly, so we adapt the permute-dim algorithm proposed in \cite{kim2018disentangling} to do the sampling, where the samples are drawn by randomly permuting across the batch for each latent dimension. Therefore, the input of our framework should be two randomly selected batches of latent variables, one of which is used to calculate the mutual information loss $\mathcal{L}_{I}$ and sample from $q(\hat{z})$, another batch is used to sample from $\prod_{i}q(\hat{z}_{i})$ in order to estimate the total correlation loss $\mathcal{L}_{TC}$.

\begin{algorithm}
    \LinesNumberedHidden
    \caption{Pseudocode of TC-GAN}
    \label{alg:psuedocode_drt}
    \KwInput{dataset $\left\{x_{k}\right\}_{k=1}^{N}$, batch size $M$, network structure Generator $G$, Discriminator $D$, Auxiliary Network $Q$, TC Discriminator $TCD$}
    \While{not converged}{
        Sample latent code $z_{1}, z_{2} \sim p(z_{a})$ \\
        Sample $x \sim p_{real}(x)$ \\
        
        $x_{fake1}=G(z_{1})$ \\
        Update $D$ by ascending $\mathcal{L}_{GAN}$ \\
        
        $\mu_{1}, \sigma_{1} = Q(x_{fake1})$ \\
        $\hat{z}_{1}=Reparameterization(\mu_{1}, \sigma_{1})$ \\
        $\mathcal{L}_{I} = NLL\char`_Loss(z_{1}, \mu_{1}, \sigma_{1})$ \\
        $\mathcal{L}_{TC} = log(TCD(\hat{z}_{1})/(1-TCD(\hat{z}_{1})))$ \\
        Update $G$, $Q$ by descending $\mathcal{L}_{TCGAN}$\\
        
        $x_{fake2}=G(z_{2})$ \\
        $\mu_{2}, \sigma_{2} = Q(x_{fake2})$ \\
        $\hat{z}_{2}=Reparameterization(\mu_{2}, \sigma_{2})$ \\
        $\hat{z}^{'}_{2}=permute$\textunderscore$dim(\hat{z}_{2})$ \\
        $\mathcal{L}_{TCD} = CrossEntropy(\hat{z}_{1}, \hat{z}^{'}_{2})$ \\
        Update $TCD$ by descending $\mathcal{L}_{TCD}$ \\
      }
     \KwOutput{Trained networks}
\end{algorithm}

We perform three steps iteratively to train the framework: 1) Train the Discriminator $D$ in GAN to ensure the generated images look real with other modules fixed; 2) Train the Generator $G$ and Network $Q$ by the loss defined in Equation~\ref{eq:loss_infotcgan} with other modules fixed, where $\mathcal{L}_{TC}$ is estimated by Equation~\ref{eq:density_ratio_trick}; 3) Train the Total Correlation Discriminator $TCD$ to distinguish $q(\hat{z})$ and $\prod_{i}q(\hat{z}_{i})$ with other modules fixed. These steps are illustrated in Algorithm~\ref{alg:psuedocode_drt}.

\section{Experiments}
\subsection{Experiment Setting}
\label{exp_setting}
We evaluate the performance of our method from two perspectives: (1) Quantitative Evaluation (Section~\ref{exp_quan_eval}), where we compare our method with several baseline methods on multiple commonly used metrics for Disentangled Representation Learning; and (2) Qualitative Evaluation (Section~\ref{exp_traversal}), where we conduct Latent Space Traversal Test and compare our results with the case without the enforcement of independence constraint, to observe the direct impact of our method on the generated images.

We follow the settings in previous works on the model architectures of the Generator, Discriminator \cite{lin2020infogan} and the Total Correlation Discriminator \cite{kim2018disentangling}. We use the Adam optimizer for training, with a learning rate equal to 0.001 for the generator, 0.002 for the discriminator and TC discriminator. We use $\alpha=0.1$ and $\beta=0.001$ in Equation~\ref{eq:loss_infotcgan}. The latent dimension equals 10. We used a batch size of 64 and trained the model for 30 epochs, then selected the checkpoint with the highest Explicitness score for evaluation on other metrics.

\subsection{Quantitative Evaluation}
\label{exp_quan_eval}

\begin{table}[t]
% \vspace{-10pt}
  \centering
  % \ssmall
  \setlength{\tabcolsep}{3pt} % Default value: 6pt
  \caption{Quantitative Evaluation of Disentanglement on dSprites. We highlight the \textbf{highest} and \underline{second highest} score for each metric in the table. We report the average and standard deviation over 10 runs for all results. }
  \begin{tabular}{l | c | c | c | c | c}
    \toprule
    & EXP & JEMMIG & MOD & SAP & Z-diff \\
    \midrule
    VAE \cite{kingma2013auto} & 0.42 $\pm$ .00 & 0.20 $\pm$ .00 & 0.87 $\pm$ .01 & 0.11 $\pm$ .00 & 0.69 $\pm$ .03 \\
    $\beta$-VAE \cite{higgins2017beta} & 0.49 $\pm$ .00 & 0.26 $\pm$ .00 & 0.82 $\pm$ .01 & 0.16 $\pm$ .00 & 0.86 $\pm$ .01 \\
    AnnealedVAE \cite{burgess2018understanding} & 0.72 $\pm$ .01 & 0.33 $\pm$ .00 & \underline{0.97 $\pm$ .00} & 0.39 $\pm$ .01 & 0.86 $\pm$ .05 \\
    Factor-VAE \cite{kim2018disentangling} & 0.41 $\pm$ .00 & 0.19 $\pm$ .00 & 0.92 $\pm$ .01 & 0.21 $\pm$ .00 & 0.80 $\pm$ .02 \\
    $\beta$-TCVAE \cite{chen2018isolating} & 0.68 $\pm$ .01 & 0.12 $\pm$ .00 & 0.90 $\pm$ .00 & 0.22 $\pm$ .00 & 0.87 $\pm$ .03 \\
    InfoGAN \cite{chen2016infogan} & 0.54 $\pm$ .00 & 0.08 $\pm$ .00 & 0.56 $\pm$ .02 & 0.05 $\pm$ .00 & 0.76 $\pm$ .04 \\
    IB-GAN \cite{jeon2021ib} & \underline{0.78 $\pm$ .02} & 0.02 $\pm$ .01 & 0.86 $\pm$ .03 & 0.19 $\pm$ .01 & 0.84 $\pm$ .04 \\
    InfoGAN-CR \cite{lin2020infogan} & 0.62 $\pm$ .00 & \underline{0.38 $\pm$ .00} & 0.95 $\pm$ .00 & \underline{0.41 $\pm$ .00} & \underline{0.99 $\pm$ .02} \\
    \midrule
    TC-GAN (ours) & \textbf{0.85 $\pm$ .01} & \textbf{0.45 $\pm$ .00} & \textbf{0.98 $\pm$ .00} & \textbf{0.48 $\pm$ .00} & \textbf{0.99 $\pm$ .01} \\
    \bottomrule
  \end{tabular}
  \label{tab:rst}
  % \vspace{-10pt}
\end{table}

Since the quantitative evaluation of disentanglement requires \textbf{ground-truth labels} of all generative factors, most datasets are not suitable for quantitative evaluation. Therefore, following previous works in this domain, we concentrate on dSprites \cite{dsprites17} for quantitative evaluation, where all factors are well-defined.

\subsubsection{Metrics}
We evaluate the quality of disentanglement by several metrics proposed in recent literature, including 1) \textbf{Explicitness Score} \cite{ridgeway2018learning} which evaluates the quality of disentanglement by training a classifier on latent code to predict the ground-truth factor classes, a higher Explicitness Score indicates that all the generative factors are decoded in the representation; 2) \textbf{JEMMIG Score} \cite{do2019theory} which evaluates the quality of disentanglement by estimating the mutual information (MI) between the ground-truth generative factors and the latent variables; 3) \textbf{Modularity Score} \cite{ridgeway2018learning} which measures whether one latent variable encodes no more than one generative factor, it estimates the mutual information (MI) between a certain latent variable and the factor with maximum MI and compares it with all other factors; 4) \textbf{SAP Score} \cite{kumar2017variational}, which evaluates the quality of disentanglement by training a linear regression model for every pair of latent variables and ground-truth factor, and uses the $R^{2}$ score of the regression model to denote the disentanglement score; and 5) \textbf{Z-diff} \cite{higgins2017beta} which first selects pairs of data points with the same value on a fixed latent variable, and then evaluates the quality of disentanglement by training a classifier to predict which factor was fixed. We used the code provided by \cite{carbonneau2022measuring} for all evaluation metrics.

\subsubsection{Baselines}
We compare our method with several baseline methods in the domain of Disentangled Representation Learning, which include VAE \cite{kingma2013auto}, $\beta$-VAE \cite{higgins2017beta}, AnnealedVAE \cite{burgess2018understanding}, Factor-VAE \cite{kim2018disentangling}, $\beta$-TCVAE \cite{chen2018isolating}, InfoGAN \cite{chen2016infogan}, IB-GAN \cite{jeon2021ib}, and InfoGAN-CR \cite{lin2020infogan}. For the VAE-based baseline methods \cite{kingma2013auto,higgins2017beta,burgess2018understanding,kim2018disentangling,chen2018isolating}, we reproduce the results by the code provided in \cite{dubois2019dvae} with the suggested optimized parameters. In particular, we use a batch size of 64 for all experiments and train the model for 30 epochs then select the checkpoint with the highest Explicitness score for evaluation. We use the model architecture suggested in \cite{burgess2018understanding} and use a latent dimension equal to 10 for all methods. For $\beta$-VAE \cite{higgins2017beta}, we train the model with $\beta$ equal to 4. For Annealed VAE \cite{burgess2018understanding}, we set the capacity equal to 25. For factor VAE and $\beta$-TCVAE, we use the weight of 6.4 for the total correlation term. On the other hand, for baseline methods based on GAN \cite{chen2016infogan,lin2020infogan,jeon2021ib}, we reproduce the result with the exact parameters and model architecture provided in the corresponding paper and code.

\subsubsection{Results} The results are presented in Table~\ref{tab:rst}, where we present
the average and standard deviation over 10 runs for all metrics, and highlight the highest score in bold, and the second-highest score by an underscore. Our method outperforms the existing methods from two perspectives:

1) Our method achieves the best performance across all the evaluation metrics. Specifically, our method outperforms baseline methods by a considerable margin on Explicitness, Modularity and SAP scores. And on the other two metrics JEMMIG and Z-diff scores, even though existing methods already performed well, our method also outperforms baseline methods by a reasonable margin. 

2) Our method exhibits consistency across the evaluation metrics. Other methods, in contrast, often lack this consistency. For instance, IB-GAN performs well on Explicitness and Modularity Score but performs poorly on JEMMIG and SAP. This inconsistency arises from the fact that different metrics perform their evaluation from different perspectives as introduced in Section~\ref{intro} and the \textit{Metrics} section above. Therefore, the consistency of our method shows the effectiveness of our method in enhancing the quality and generalizability of Disentangled Representation Learning algorithms.

\subsection{Qualitative Evaluation by Latent Space Traversal Test}
\label{exp_traversal}
While the datasets suitable for quantitative evaluation are limited, we further evaluate our method on other datasets by conducting Latent Space Traversal Tests. \textbf{Latent space traversal test} is a commonly used technique to investigate the semantic meaning of latent variables. It traverses one latent variable while keeping all the other variables invariant, and generates a sequence of images with these features. The semantic meaning of the traversed variable can be obtained by inspecting the changes in the images.

In our experiments, since we aim to improve the quality of disentanglement, we examine whether the changes in one variable affect more than one generative factor. If the traversed variable affects \textbf{only one} semantic meaningful factor, it means the quality of disentanglement is desirable. In contrast, if the traversed variable affects \textbf{more than one} semantic meaningful factor, it means the factors are still entangled in the latent space.

To this end, we make comparisons between the cases with/without the enforcement of the proposed independence constraint and directly observe the impact of implementing our method. To ensure fair comparisons, all the settings remain the same except for the total correlation loss $\mathcal{L}_{TC}$ (Equation~\ref{eq:loss_tcloss}) in each pair of comparisons. These comparisons are conducted on three datasets: MNIST, FashionMNIST and dSprites. For all three datasets, the settings remain unchanged as introduced in Section~\ref{exp_setting}, except for the dimension settings of the latent space, because the number of generative factors contained in each dataset is different. This will be further elaborated in each paragraph below.

\begin{figure}[t]
\centering
  \begin{subfigure}[b]{0.45\textwidth}
        \includegraphics[width=\linewidth]{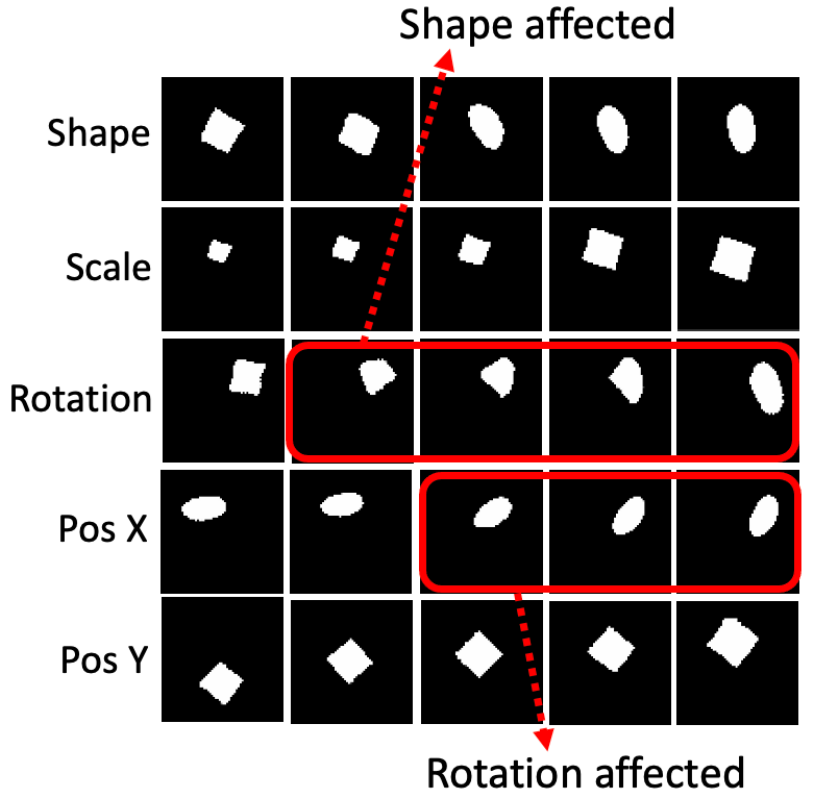}
        \caption{}
        \label{fig:dsprites_samples_without}
    \end{subfigure}%
    \begin{subfigure}[b]{0.45\textwidth}
        \includegraphics[width=\linewidth]{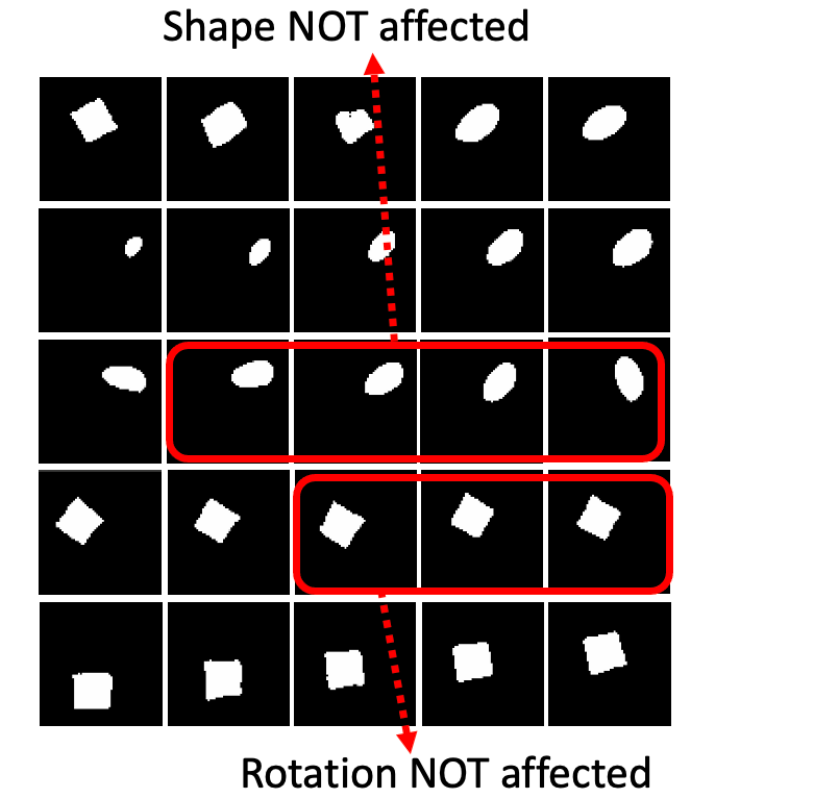}
        \caption{}
        \label{fig:dsprites_samples_with}
    \end{subfigure}
  \caption{dSprites Samples for comparison. Both models are trained with 5 continuous variables, and 5 noise variables. (a) Without independence constraint, the shape is affected by traversing on rotation and Rotation is affected by traversing on PosX; (b) With independence constraint, these factors are unaffected by traversing on another variable.}
  \label{fig:dsprites_samples}
\end{figure}

\subsubsection{MNIST} 
We trained the models with 1 ten-dimensional discrete variable, 2 continuous variables, and 62 noise variables. The ideal outcome of Disentangled Representation Learning is that the data encoder will map the discrete variable to the \textit{digit class}, and the two continuous variables will correspond to the \textit{width} and \textit{rotation} of the digit. The results of our experiment are provided in Figure~\ref{fig:mnist_samples}. In each row, we keep the discrete variable \textit{digit} invariant and traverse on the continuous variable \textit{rotation}. When the model is trained without the independence constraint (Figure~\ref{fig:mnist_samples_without}), we observe that: 1) The variables \textit{digit class} and \textit{rotation} are not fully disentangled in the latent space. For example, while we control the \textit{digit} to be 5 on the fifth row, some 0 and 6 are generated when traversing on \textit{rotation}; and 2) the variables \textit{width} and \textit{rotation} of the digit are not fully disentangled in the latent space. For example, as highlighted in the first row of Figure~\ref{fig:mnist_samples_without}, while we only traverse the variable \textit{rotation}, the \textit{width} of the digit is also affected. We highlight this trend on the first row, and similar patterns can be observed on other rows. In contrast, with the enhancement of the independent constraint (Figure~\ref{fig:mnist_samples_with}), both \textit{digit} and \textit{width} remain unchanged when traversing on \textit{rotation}.

\subsubsection{dSprites} 
We trained the models with 5 continuous variables, and 5 noise variables. Ideally, the five continuous variables will be mapped to the five generative factors of the dataset, which include \textit{Shape}, \textit{Scale}, \textit{Rotation}, \textit{Pos X} and \textit{Pos Y}. The results of our experiment are provided in Figure~\ref{fig:dsprites_samples}. On each row of the images, we traverse one factor while keeping all other factors invariant as noted on the left side of each row. When the model is trained without the independence constraint (Figure~\ref{fig:dsprites_samples_without}), we observe that the factors are entangled in the latent space. For example, on the third row of the image, while traversing the factor of \textit{rotation}, the \textit{shape} of the figures are affected. And on the fourth row, \textit{rotation} is affected while traversing the \textit{Position X}. In contrast, with the enhancement of the independent constraint (Figure~\ref{fig:dsprites_samples_with}), these factors are not affected, as highlighted by red boxes in Figure~\ref{fig:dsprites_samples}.

\begin{figure}[t]
\centering
  \begin{subfigure}[b]{0.48\textwidth}
        \includegraphics[width=\linewidth]{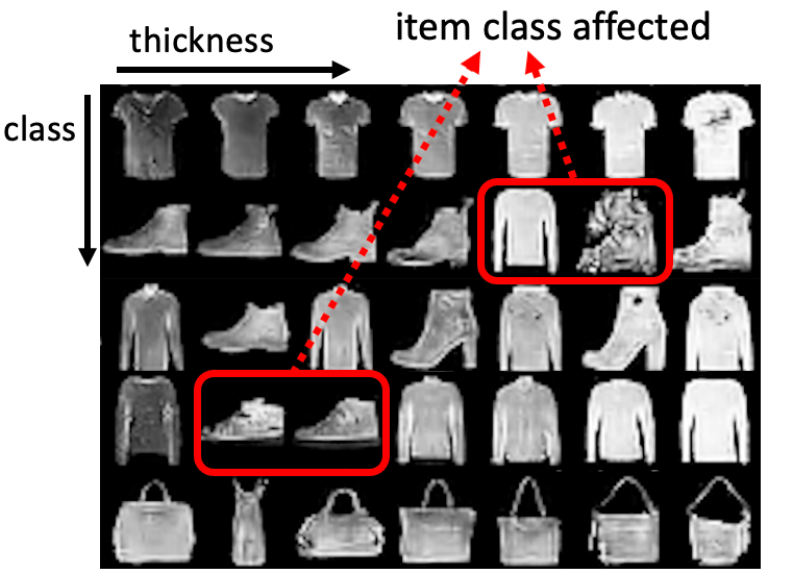}
        \caption{}
        \label{fig:fmnist_samples_without}
    \end{subfigure}%
    \begin{subfigure}[b]{0.48\textwidth}
        \includegraphics[width=\linewidth]{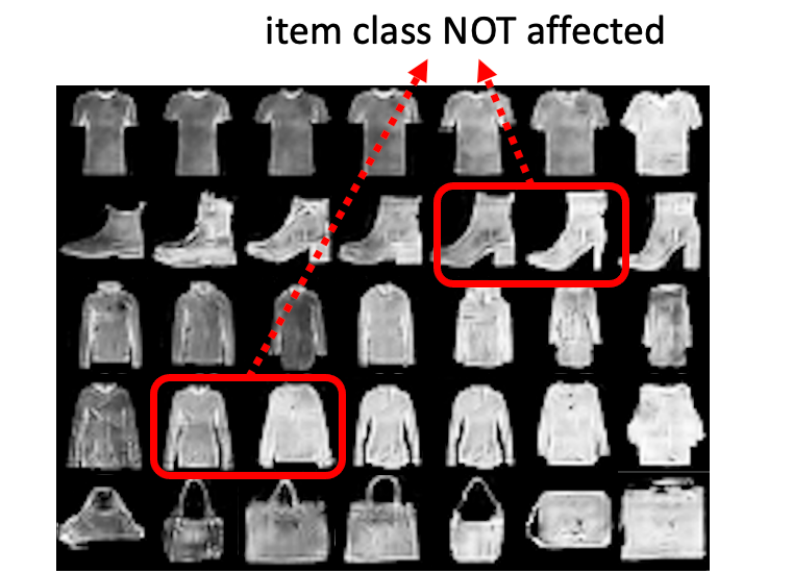}
        \caption{}
        \label{fig:fmnist_samples_with}
    \end{subfigure}
  \caption{FashionMNIST Samples for comparison. Both models are trained with 1 ten-dimensional discrete variable, 1 continuous variable, and 62 noise variables. (a) Without constraining on the independence between variables, item type is affected by traversing on thickness. (b) With the constraint on the independence between variables, item type is unaffected while traversing on thickness.}
  \label{fig:fmnist_samples}
  % \vspace{-20pt}
\end{figure}

\subsubsection{FashionMNIST} 
We trained the models with 1 ten-dimensional discrete variable, 1 continuous variable, and 62 noise variables. Ideally, the discrete variable and the continuous variable will correspond to the \textit{item class} and the \textit{thickness} of the image. The results of our experiment are provided in Figure~\ref{fig:fmnist_samples}. A similar conclusion can be drawn as we did on other datasets above. On each row, we keep the discrete variable \textit{item class} invariant and traverse on the continuous variable \textit{thickness} of the image. When the model is trained without the constraint of the independence between latent variables (Figure~\ref{fig:fmnist_samples_without}), we observe that the item is affected by variable \textit{thickness}, as highlighted on the second row and the fourth row. In contrast, the item type remains unaffected when the model is trained with the enhancement of the independent constraint (Figure~\ref{fig:fmnist_samples_with}).

\subsubsection{Summary} Based on these comparisons, we conclude that our method could consistently enhance the quality of disentanglement. Note that this does not mean our method could always disentangle all the factors perfectly, however, while some factors may still exhibit slight entanglement in the given images, the differences between the cases with/without the independence constraint are nontrivial, which validates the effectiveness of our method.

\section{Related Works}
\subsection{Disentangled Representation Learning}
Disentangled Representation Learning aims to learn a data encoder that can identify true latent variables that are semantically meaningful. \cite{higgins2017beta} suggested that increasing the weight of the KL regularizer in VAE \cite{kingma2013auto} can benefit the quality of disentangled representation learning. \cite{burgess2018understanding} proposed that disentanglement quality can be improved by progressively increasing the bottleneck capacity. FactorVAE \cite{kim2018disentangling} and $\beta$-TCVAE \cite{chen2018isolating} both penalize the total correlation \cite{watanabe1960information} between latent variables, while FactorVAE uses a density-ratio trick for total correlation estimation, $\beta$-TCVAE proposed a biased Monte-Carlo estimator to approximate total correlation. Several other methods were also proposed in the paradigm of VAE \cite{kumar2017variational,dupont2018learning,kim2019relevance}. However, \cite{locatello2019challenging} claimed that unsupervised disentangled representation learning is impossible without inductive biases. 

On the other hand, some methods are proposed to learn disentangled representation in the paradigm of GAN \cite{goodfellow2020generative,zhu2021and}. InfoGAN \cite{chen2016infogan} proposed a method to learn disentangled representation by maximizing the mutual information between latent variables and the generated image. InfoGAN-CR \cite{lin2020infogan} claimed that self-supervision techniques can be used to improve the quality of disentanglement. IB-GAN \cite{jeon2021ib} utilized the Information Bottleneck framework for the optimization of GAN. Besides, some methods attempted to learn disentangled representations without utilizing generative models\cite{wang2021self}.

Recently, diffusion models have been applied to the domain of disentangled representation learning \cite{yang2023disdiff,chen2023disenbooth,chen2024disendreamer}. Additionally, disentangled representation learning has found broad applications in areas such as graph representation learning \cite{li2024disentangled}, graph neural architecture search \cite{zhangdisentangled}, recommendation systems \cite{zhang2023adaptive,wang2023curriculum}, and out-of-distribution generalization \cite{yoo2023disentangling}.

\subsection{Causal Representation Learning}
Recent studies are aimed at connecting the field of causal inference and disentangled representation learning. \cite{suter2019robustly} introduced a causal perspective of disentangled representation learning by modelling the data generation process as a Structural Causal Model (SCM) \cite{pearl2009causality}, where they introduced a set of confounders that causally influence the generative factors of observable data. \cite{reddy2022causally} further developed this idea and studied the role of intervention and counterfactual effects. CausalVAE \cite{yang2021causalvae} introduced a fully supervised method that builds a Causal Layer to transform independent exogenous factors into causal endogenous factors that correspond to causally related concepts in the observed data. And DEAR \cite{shen2022weakly} proposed a weakly supervised framework, which learns causally disentangled representation with SCM as prior.

\section{Conclusion}
In this paper, we investigated the prior disagreement on the interrelationships between latent variables in Disentangled Representation Learning and proposed a novel method to improve the quality of disentanglement. First, we build a conceptual bridge between epistemology and disentangled representation learning, thus clarifying what should and should not be independent in the latent space by introducing a two-level latent space framework based on interdisciplinary theories. Then, after clarifying the legitimacy of applying the independence constraint on the problem of Disentangled Representation Learning, we introduce a novel method that applies the mutual information constraint and independence constraint within the Generative Adversarial Network (GAN) framework. Experiments show that our method consistently achieves better disentanglement performance on multiple evaluation metrics, and Qualitative Evaluation results show that our method leads to an improved quality for controllable generation. Besides, our paper introduced a novel perspective to apply causal models to the field of representation learning, which facilitates the development of explainability of deep learning and holds potential for wide-ranging applications that value explainability, transparency and controllability.

\bibliographystyle{splncs04}
\bibliography{6875}

\begin{thebibliography}{10}
\providecommand{\url}[1]{\texttt{#1}}
\providecommand{\urlprefix}{URL }
\providecommand{\doi}[1]{https://doi.org/#1}

\bibitem{bengio2013representation}
Bengio, Y., Courville, A., Vincent, P.: Representation learning: A review and new perspectives. IEEE transactions on pattern analysis and machine intelligence  \textbf{35}(8),  1798--1828 (2013)

\bibitem{berlin2014ethnobiological}
Berlin, B.: Ethnobiological classification: Principles of categorization of plants and animals in traditional societies, vol.~185. Princeton University Press (2014)

\bibitem{burgess2018understanding}
Burgess, C.P., Higgins, I., Pal, A., Matthey, L., Watters, N., Desjardins, G., Lerchner, A.: Understanding disentangling in $beta $-vae. arXiv preprint arXiv:1804.03599  (2018)

\bibitem{carbonneau2022measuring}
Carbonneau, M.A., Zaidi, J., Boilard, J., Gagnon, G.: Measuring disentanglement: A review of metrics. IEEE Transactions on Neural Networks and Learning Systems  (2022)

\bibitem{chen2023disenbooth}
Chen, H., Zhang, Y., Wang, X., Duan, X., Zhou, Y., Zhu, W.: Disenbooth: Disentangled parameter-efficient tuning for subject-driven text-to-image generation. arXiv preprint arXiv:2305.03374  \textbf{3} (2023)

\bibitem{chen2024disendreamer}
Chen, H., Zhang, Y., Wang, X., Duan, X., Zhou, Y., Zhu, W.: Disendreamer: Subject-driven text-to-image generation with sample-aware disentangled tuning. IEEE Transactions on Circuits and Systems for Video Technology  (2024)

\bibitem{chen2018isolating}
Chen, R.T., Li, X., Grosse, R.B., Duvenaud, D.K.: Isolating sources of disentanglement in variational autoencoders. Advances in neural information processing systems  \textbf{31} (2018)

\bibitem{chen2016infogan}
Chen, X., Duan, Y., Houthooft, R., Schulman, J., Sutskever, I., Abbeel, P.: Infogan: Interpretable representation learning by information maximizing generative adversarial nets. Advances in neural information processing systems  \textbf{29} (2016)

\bibitem{do2019theory}
Do, K., Tran, T.: Theory and evaluation metrics for learning disentangled representations. arXiv preprint arXiv:1908.09961  (2019)

\bibitem{dubois2019dvae}
Dubois, Y., Kastanos, A., Lines, D., Melman, B.: Disentangling vae. \url{http://github.com/YannDubs/disentangling-vae/} (march 2019)

\bibitem{dupont2018learning}
Dupont, E.: Learning disentangled joint continuous and discrete representations. Advances in Neural Information Processing Systems  \textbf{31} (2018)

\bibitem{goodfellow2020generative}
Goodfellow, I., Pouget-Abadie, J., Mirza, M., Xu, B., Warde-Farley, D., Ozair, S., Courville, A., Bengio, Y.: Generative adversarial networks. Communications of the ACM  \textbf{63}(11),  139--144 (2020)

\bibitem{higgins2018towards}
Higgins, I., Amos, D., Pfau, D., Racaniere, S., Matthey, L., Rezende, D., Lerchner, A.: Towards a definition of disentangled representations. arXiv preprint arXiv:1812.02230  (2018)

\bibitem{higgins2017beta}
Higgins, I., Matthey, L., Pal, A., Burgess, C., Glorot, X., Botvinick, M., Mohamed, S., Lerchner, A.: beta-vae: Learning basic visual concepts with a constrained variational framework. In: International conference on learning representations (2017)

\bibitem{hume2003treatise}
Hume, D.: A treatise of human nature. Courier Corporation (2003)

\bibitem{jeon2021ib}
Jeon, I., Lee, W., Pyeon, M., Kim, G.: Ib-gan: Disentangled representation learning with information bottleneck generative adversarial networks. In: Proceedings of the AAAI Conference on Artificial Intelligence. vol.~35, pp. 7926--7934 (2021)

\bibitem{kim2018disentangling}
Kim, H., Mnih, A.: Disentangling by factorising. In: International Conference on Machine Learning. pp. 2649--2658. PMLR (2018)

\bibitem{kim2019relevance}
Kim, M., Wang, Y., Sahu, P., Pavlovic, V.: Relevance factor vae: Learning and identifying disentangled factors. arXiv preprint arXiv:1902.01568  (2019)

\bibitem{kingma2013auto}
Kingma, D.P., Welling, M.: Auto-encoding variational bayes. arXiv preprint arXiv:1312.6114  (2013)

\bibitem{kumar2017variational}
Kumar, A., Sattigeri, P., Balakrishnan, A.: Variational inference of disentangled latent concepts from unlabeled observations. arXiv preprint arXiv:1711.00848  (2017)

\bibitem{li2024disentangled}
Li, H., Wang, X., Zhang, Z., Chen, H., Zhang, Z., Zhu, W.: Disentangled graph self-supervised learning for out-of-distribution generalization. In: Forty-first International Conference on Machine Learning (2024)

\bibitem{lin2020infogan}
Lin, Z., Thekumparampil, K., Fanti, G., Oh, S.: Infogan-cr and modelcentrality: Self-supervised model training and selection for disentangling gans. In: international conference on machine learning. pp. 6127--6139. PMLR (2020)

\bibitem{locatello2019challenging}
Locatello, F., Bauer, S., Lucic, M., Raetsch, G., Gelly, S., Sch{\"o}lkopf, B., Bachem, O.: Challenging common assumptions in the unsupervised learning of disentangled representations. In: international conference on machine learning. pp. 4114--4124. PMLR (2019)

\bibitem{dsprites17}
Matthey, L., Higgins, I., Hassabis, D., Lerchner, A.: dsprites: Disentanglement testing sprites dataset. https://github.com/deepmind/dsprites-dataset/ (2017)

\bibitem{murphy2004big}
Murphy, G.: The big book of concepts. MIT press (2004)

\bibitem{pearl2009causality}
Pearl, J.: Causality. Cambridge university press (2009)

\bibitem{reddy2022causally}
Reddy, A.G., Balasubramanian, V.N., et~al.: On causally disentangled representations. In: Proceedings of the AAAI Conference on Artificial Intelligence. vol.~36, pp. 8089--8097 (2022)

\bibitem{ridgeway2018learning}
Ridgeway, K., Mozer, M.C.: Learning deep disentangled embeddings with the f-statistic loss. Advances in neural information processing systems  \textbf{31} (2018)

\bibitem{rips1973semantic}
Rips, L.J., Shoben, E.J., Smith, E.E.: Semantic distance and the verification of semantic relations. Journal of verbal learning and verbal behavior  \textbf{12}(1),  1--20 (1973)

\bibitem{shen2022weakly}
Shen, X., Liu, F., Dong, H., Lian, Q., Chen, Z., Zhang, T.: Weakly supervised disentangled generative causal representation learning. Journal of Machine Learning Research  \textbf{23},  1--55 (2022)

\bibitem{suter2019robustly}
Suter, R., Miladinovic, D., Sch{\"o}lkopf, B., Bauer, S.: Robustly disentangled causal mechanisms: Validating deep representations for interventional robustness. In: International Conference on Machine Learning. pp. 6056--6065. PMLR (2019)

\bibitem{wang2021self}
Wang, T., Yue, Z., Huang, J., Sun, Q., Zhang, H.: Self-supervised learning disentangled group representation as feature. Advances in Neural Information Processing Systems  \textbf{34},  18225--18240 (2021)

\bibitem{wang2023curriculum}
Wang, X., Pan, Z., Zhou, Y., Chen, H., Ge, C., Zhu, W.: Curriculum co-disentangled representation learning across multiple environments for social recommendation. In: International Conference on Machine Learning. pp. 36174--36192. PMLR (2023)

\bibitem{watanabe1960information}
Watanabe, S.: Information theoretical analysis of multivariate correlation. IBM Journal of research and development  \textbf{4}(1),  66--82 (1960)

\bibitem{yang2021causalvae}
Yang, M., Liu, F., Chen, Z., Shen, X., Hao, J., Wang, J.: Causalvae: Disentangled representation learning via neural structural causal models. In: Proceedings of the IEEE/CVF conference on computer vision and pattern recognition. pp. 9593--9602 (2021)

\bibitem{yang2023disdiff}
Yang, T., Wang, Y., Lv, Y., Zheng, N.: Disdiff: Unsupervised disentanglement of diffusion probabilistic models. arXiv preprint arXiv:2301.13721  (2023)

\bibitem{yoo2023disentangling}
Yoo, H., Lee, Y.C., Shin, K., Kim, S.W.: Disentangling degree-related biases and interest for out-of-distribution generalized directed network embedding. In: Proceedings of the ACM Web Conference 2023. pp. 231--239 (2023)

\bibitem{zhang2023adaptive}
Zhang, Y., Wang, X., Chen, H., Zhu, W.: Adaptive disentangled transformer for sequential recommendation. In: Proceedings of the 29th ACM SIGKDD Conference on Knowledge Discovery and Data Mining. pp. 3434--3445 (2023)

\bibitem{zhangdisentangled}
Zhang, Z., Wang, X., Qin, Y., Chen, H., Zhang, Z., Chu, X., Zhu, W.: Disentangled continual graph neural architecture search with invariant modular supernet. In: Forty-first International Conference on Machine Learning

\bibitem{zhu2021and}
Zhu, X., Xu, C., Tao, D.: Where and what? examining interpretable disentangled representations. In: Proceedings of the IEEE/CVF Conference on Computer Vision and Pattern Recognition. pp. 5861--5870 (2021)

\end{thebibliography}

\end{document}